\documentclass[journal,transmag]{IEEEtran}

\usepackage{amsmath,amssymb}
\usepackage{booktabs}
\usepackage{cite}
\usepackage{graphicx}
\usepackage{url}
\usepackage{xcolor}
\usepackage{microtype}
\usepackage{placeins}

\graphicspath{{figures/}}
\DeclareMathOperator*{\argmax}{arg\,max}

\begin{document}

\title{Baseline-Relative Counterfactual Refinement for Bit-Aware Visual Token Communication}

\author{Jia Guo, ~\IEEEmembership{Member,~IEEE},  Xiaohan Zhao, Changwang Liu*, ~\IEEEmembership{Member,~IEEE}, \\ Shuqing He, Chenyang Zhang, Bingchuan Zhao, Jinqi Zhu

 \thanks{This work was supported by the National Natural Science Foundation of China under Grant No.~62002263, the Tianjin Science and Technology Program Projects under Grant No.~24YDTPJC00630, and the Tianjin Research Innovation Project for Postgraduate Students under Grant No.~2026KYCX001F.}
 \thanks{J. Guo, X.Zhao, C. Liu, S.He, C. Zhang, B. Zhao and J. Zhu are with the School of Computer and Information Engineering, Tianjin Normal University, Tianjin, China (e-mail: c04s316@bupt.cn, 2411090059@stu.tjnu.edu.cn, 004967@tjnu.edu.cn, 2411090038@stu.tjnu.edu.cn, 2411090077@stu.tjnu.edu.cn,jsjzhujinqi@tjnu.edu.cn}
 \thanks{S. He is with the School of Information Science and Engineering, Linyi University, Linyi 276000, China (email:heshuqing@lyu.edu.cn)}
 \thanks{*Corresponding Author: Changwang Liu} 
  \thanks{The source code for this work is publicly available at: https://github.com/c04s316/GCR-C}
}

\IEEEtitleabstractindextext{%
\begin{abstract}
Generative visual-token communication reduces transmission load by sending only selected discrete tokens and reconstructing missing content at the receiver. However, existing token-selection criteria based on local uncertainty, importance, or diversity do not directly determine whether changing the current selection improves the final reconstruction under the same packet budget. To address this problem, we propose Gated Counterfactual Refinement for Communication (GCR-C), a rollout-style correction layer over Local-MDL. GCR-C constructs a compact diversified candidate set, evaluates each candidate through matched full-budget Local-MDL continuation, and replaces the baseline action only when a positive baseline-relative reconstruction gain is obtained. Experiments on CIFAR-10, STL-10, a coded 5G-LDPC link, and a limited high-resolution Kodak transfer show that GCR-C consistently improves reconstruction quality at active low- and medium-rate operating points without increasing the realized packet rate, while remaining effective across changes in dataset, channel condition, resolution, token grid, and tokenizer. The results also reveal a clear quality--computation tradeoff due to the additional encoder-side counterfactual evaluation.
\end{abstract}
\begin{IEEEkeywords}
Semantic communications, visual tokens, generative recovery, rate--distortion, rollout policy improvement.
\end{IEEEkeywords}}

\maketitle
\IEEEdisplaynontitleabstractindextext

\section{Introduction}

Recent advances in discrete visual representation and generative modeling are reshaping image communication. Vector-quantized representations convert images into discrete visual tokens, while masked generative models can recover missing tokens from partial observations \cite{oord2017neural,esser2021taming,chang2022maskgit}. Together with learned image compression, deep joint source--channel coding, and semantic communication \cite{minnen2018joint,bourtsoulatze2019deep,xie2021deep,han2022generative,pezone2025sqgan}, this enables the transmitter to send only selected visual evidence and rely on receiver-side generative recovery. Under a constrained link, reconstruction quality therefore depends not only on coding efficiency, but also on \emph{which} visual tokens are transmitted.

This paradigm is particularly relevant when communication resources are scarce but transmitter-side computation is comparatively available. Deep-space and satellite communications are representative examples: scientific instruments and high-resolution imagers can generate large data volumes, whereas long-distance propagation, limited bandwidth, constrained transmit power, and intermittent connectivity restrict return rates \cite{edwards2024deepspace}. Meanwhile, modern onboard processors can perform substantial preprocessing and data reduction before transmission \cite{lin2011spacecube}. Similar conditions arise in remote sensing and compute-capable edge platforms. In such systems, additional transmitter computation can be worthwhile if it improves the utility of the information delivered under the same packet budget, motivating a \emph{compute-for-rate} design.

The resulting problem is not merely \emph{how many} tokens to transmit, but \emph{which} tokens should consume the available budget. A transmitted token carries its own visual evidence while also providing context for recovering untransmitted tokens. Its value therefore depends on the already protected set and the subsequent completion process. Moreover, under a bit-accurate packet protocol, token positions also consume bits, so packet cost depends on the selected set rather than only on its cardinality. The key question is therefore: \emph{if the current baseline action is replaced by another feasible token, will the completed reconstruction improve under the same budget and continuation policy?}

Existing visual-token reduction methods use uncertainty, attention, redundancy, diversity, contextual relevance, or semantic importance to prioritize tokens. Progressive compression, diversity-aware pruning, clustering, representation-shift analysis, and redundancy modeling have been studied for visual-token reduction \cite{ye2025voco,yang2025pvc,alvar2025divprune,bergner2025tokencropr,dhouib2025pact,choi2025representationshift,kim2025tokenredundancy}. Adaptive, contextual, language-guided, and training-free selection methods provide additional signals \cite{zeng2025skipvision,ye2025atp,zha2025textok,ye2025fitprune,guo2025crop,li2025inference}, while communication-oriented studies have explored text-guided transmission, adaptive semantic tokens, semantic subspaces, and generative image communication \cite{liu2025textguided,devoto2025adaptive,li2026unisc,pezone2025sqgan}.

These methods provide efficient selection cues, but they do not directly measure the \emph{baseline-relative terminal reconstruction value} of a feasible action under the same packet budget. A high-scoring token may be redundant with already transmitted evidence, whereas a lower-scoring but spatially complementary token may improve generative recovery. More importantly, an action that appears better locally may become inferior after the remaining budget is completed. Instantaneous scores are therefore only proxies for the final reconstruction objective.

Directly optimizing the final reconstruction is difficult. Searching all packet-feasible token subsets is combinatorial, while exhaustively evaluating every feasible next action with full-budget completion is also expensive. The practical challenge is therefore to identify useful deviations from a strong baseline without turning each transmission decision into exhaustive search.

To address this problem, we propose \emph{Gated Counterfactual Refinement for Communication} (GCR-C), a baseline-relative correction layer over Local-MDL. Local-MDL remains the default policy. At an eligible state, GCR-C constructs a compact diversified proposal containing high-value Local candidates and spatially complementary alternatives. Each candidate is taken once counterfactually, after which the remaining packet budget is completed using the \emph{same} Local-MDL continuation. Its full-budget reconstruction quality is then compared with that obtained from the original Local-MDL action under the identical continuation. GCR-C replaces the baseline action only when the best candidate has strictly positive baseline-relative advantage; otherwise, it falls back to Local-MDL. Rate/progress gating and a limit on accepted interventions further control encoder-side computation.

GCR-C is related to rollout-style policy improvement, where a candidate action is taken once and the remaining decisions follow a fixed base policy \cite{bertsekas2019biased}. We do not claim the generic rollout principle as new. Our contribution is its specialization to \emph{bit-accurate generative visual-token communication}, where feasibility is determined by the serialized packet, candidate value is defined by same-budget receiver-side reconstruction, and counterfactual evaluation incurs non-negligible encoder cost. This differs from world-model-based semantic communication for Physical-AI systems, which uses imagined rollouts to evaluate long-horizon control return \cite{wang2026wmcdt}; here, the objective is same-image reconstruction under a fixed receiver continuation.

The intended operating regime of GCR-C is therefore one in which \emph{communication is more constrained than transmitter-side computation}. Representative scenarios include bandwidth- or power-limited satellite imaging links, deep-space scientific-image return from compute-capable spacecraft, remote-sensing platforms, and other compute-capable edge transmitters connected through restricted links. In such systems, spending additional computation to select more useful visual evidence can be preferable to transmitting more bits. Conversely, the current full-budget evaluator is less suitable for ultra-low-power transmitters or strict real-time applications where encoder computation and latency are dominant constraints.

The main contributions of this work are summarized as follows:
\begin{enumerate}
\item We formulate a packet-feasible, baseline-relative fixed-budget correction problem for generative visual-token communication. Candidate and baseline actions are compared through matched full-budget completion under the same packet syntax, transmission budget, and Local-MDL continuation, and we establish a baseline-preservation proposition under exact deterministic matched completion.

\item We develop GCR-C, a non-learned correction layer that combines a compact diversified proposal, full-budget Local-MDL continuation, baseline-relative gating, rate/progress eligibility, Local fallback, and a limit on accepted interventions, without redesigning the tokenizer, masked prior, packet format, or receiver pipeline.

\item We introduce counterfactual headroom, proposal regret, and quality--computation diagnostics to quantify the opportunity remaining above Local-MDL, the fraction exposed by a compact proposal, and the encoder-side cost required for refinement.

\item We evaluate GCR-C across multiple rates, datasets, channel conditions, and representation scales. On a held-out 500-image CIFAR-10 test split, GCR-C improves PSNR over Local-MDL by $1.503$ dB at $0.20$ bpp and $0.698$ dB at $0.32$ bpp without increasing the realized packet rate. Frozen selections retain positive paired gains under all 12 tested 5G-LDPC/QPSK AWGN and block-Rayleigh conditions. An independently trained $96\times96$ STL-10 system and a limited $384\times384$ Kodak transfer further test the mechanism under changes in dataset, resolution, token grid, and vocabulary.

\end{enumerate}

\begin{figure*}[tbp]
\centering
\includegraphics[width=0.90\textwidth]{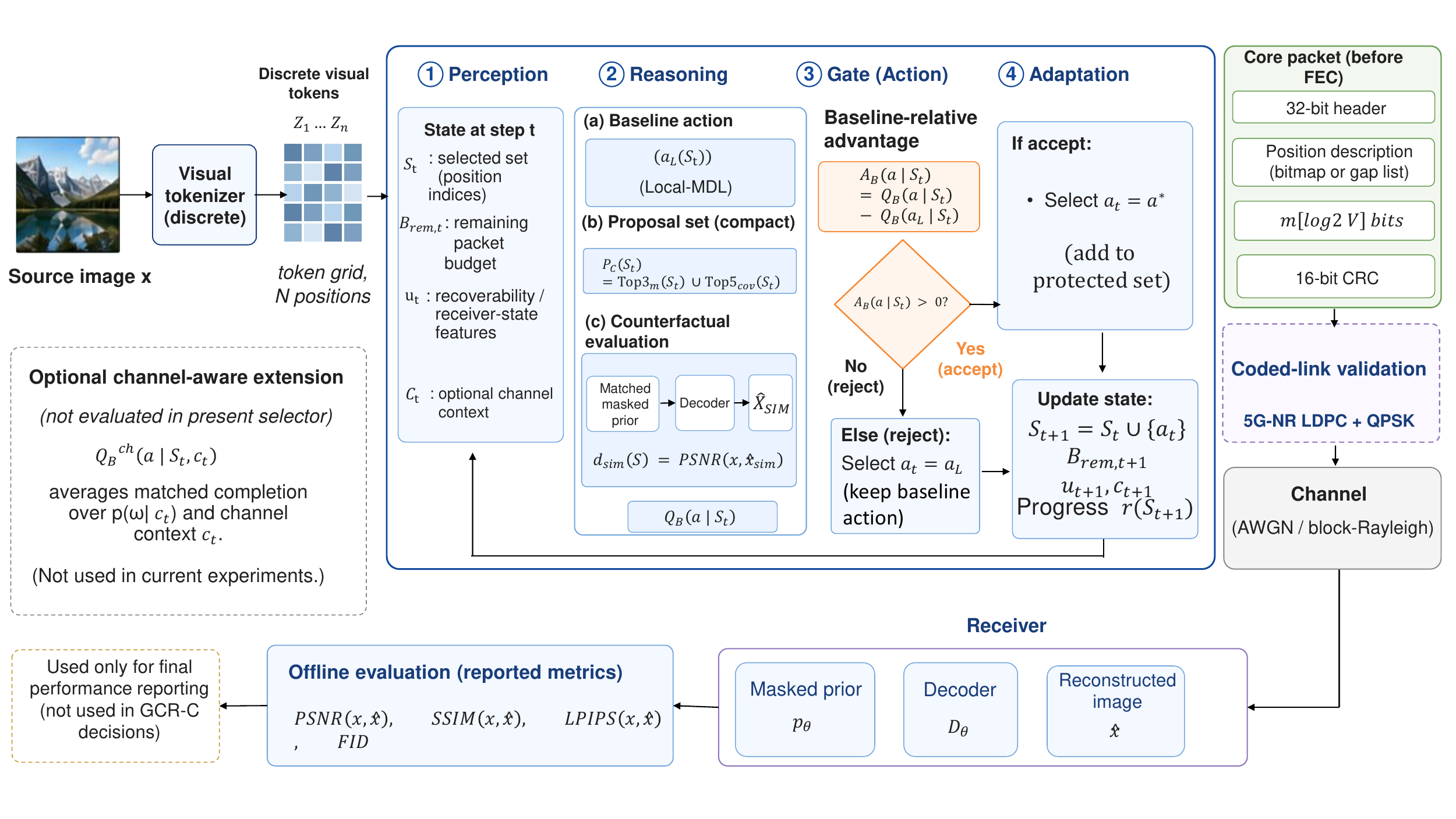}
\caption{System architecture of GCR-C. A compact proposal is evaluated by full-budget Local-MDL completion, and an alternative action is accepted only when its baseline-relative advantage is positive. Selected tokens are packetized and transmitted through the coded link; receiver reconstruction is used only for offline evaluation.}
\label{fig:cognitive-loop}
\end{figure*}

\section{Related Work}
\subsection{Discrete visual representations and generative recovery}
Vector-quantized representations make images accessible to discrete sequence models, while masked generation provides a receiver-side mechanism for recovering untransmitted tokens \cite{oord2017neural,esser2021taming,chang2022maskgit}. Learned compression and autoregressive or hierarchical priors similarly connect a rate budget with a probabilistic description of missing content \cite{minnen2018joint,rissanen1989stochastic}. In semantic communication, deep joint source--channel coding, learned semantic transmission, and generative image communication have emphasized task-aware recovery rather than pixel-perfect transport \cite{bourtsoulatze2019deep,xie2021deep,han2022generative}. These works motivate our protocol, but generally do not isolate the incremental value of a candidate set under a fixed packet syntax and a fixed completion policy.

\subsection{Visual-token reduction, contextual selection, and semantic communication}
Recent work reduces visual-token redundancy through progressive or diversity-aware compression, clustering, representation-shift analysis, and token-redundancy modeling \cite{ye2025voco,yang2025pvc,alvar2025divprune,bergner2025tokencropr,dhouib2025pact,choi2025representationshift,kim2025tokenredundancy}; token skipping and adaptive pruning have also been studied for multimodal and video systems \cite{zeng2025skipvision,ye2025atp,endo2025feather,liu2025kvtp,wang2025dynamicvlm}. Context-aware pruning, language-guided tokenization, training-free selection, and inference-optimal allocation broaden the signals available to a selector \cite{zha2025textok,ye2025fitprune,guo2025crop,li2025fcot,li2025inference}. These methods establish token count as a useful control variable, but generally target acceleration or multimodal inference rather than packet-level rate--distortion decisions with receiver-side generative completion.

Communication-oriented studies have explored text-guided transmission and adaptive semantic tokens for edge inference \cite{liu2025textguided,devoto2025adaptive}; newer directions include matchable semantic subspaces, attention-driven self-compression, reading-twice pruning, energy-aware pruning, and robustness of compressed visual tokens \cite{li2026unisc,deniz2026adsc,wan2026rtprune,chen2026energy,gu2026robust}, while wireless image-compression work exposes coupled robustness, throughput, and latency constraints \cite{wireless2026lic}. SQ-GAN combines semantic-conditioned masking with vector quantization for very-low-rate image communication \cite{pezone2025sqgan}. GCR-C is complementary: it does not redesign the tokenizer or learn a semantic mask, but uses local, set, spatial-coverage, and contextual proposals and asks whether a feasible replacement improves the completed reconstruction under the same packet budget and continuation policy.

\subsection{Rollout-style policy improvement and self-positioning}
From approximate dynamic programming, GCR-C is related to rollout-based policy improvement, where a candidate action is taken once and the remainder is completed with a known base policy \cite{bertsekas2019biased}. The rollout principle itself is therefore not claimed as new. Our specialization is to a bit-accurate generative visual-token communication problem in which feasibility depends on complete packet syntax and selected positions, exhaustive first-action evaluation is expensive, and the receiver uses generative completion. We add a compact diversified proposal, baseline-relative headroom and proposal-regret diagnostics, rate-dependent gating, explicit encoder-computation accounting, and coded-link validation. In contrast to long-horizon task return or closed-loop physical-agent planning, the present objective is same-budget image reconstruction under a fixed receiver continuation.
Recent work on physical-AI semantic communication evaluates semantic-token interventions through world-model-enabled counterfactual imagined rollouts to estimate long-horizon return-per-bit \cite{wang2026wmcdt}. That setting couples communication with future control and state evolution. GCR-C addresses a different problem: same-budget image reconstruction under a fixed receiver continuation, with Local-MDL serving as an explicit base policy and packet feasibility determined by the serialized token packet.

Prior work provides effective local, contextual, spatial, and semantic signals for reducing visual tokens, but these signals are typically optimized for pruning efficiency, downstream inference, or adaptive transmission. GCR-C addresses the narrower decision of whether a feasible replacement improves receiver reconstruction after both choices are completed under the same budget and continuation policy. We adapt the published DivPrune max--min diversity principle under our packet interface as a matched external selector \cite{alvar2025divprune}.

\section{Problem Formulation and Bit-Accurate Communication Protocol}
\subsection{Sequential baseline-correction problem}
Given a receiver reconstruction function $d(S)$ and a transmission budget $B_{\max}$, the communication objective is to construct a feasible transmitted set with high completed reconstruction quality subject to $B(S)\leq B_{\max}$. Direct global subset search is combinatorial and is not attempted here. Instead, we study a sequential decision problem in which a default Local-MDL action is available at each state and ask whether a feasible alternative can improve the same-budget completed reconstruction.

\subsection{Tokenized image and packet budget}
Let an image $x\in[0,1]^{H\times W\times3}$ be divided into $N$ non-overlapping patches and quantized into tokens $z=(z_1,\ldots,z_N)$ from a vocabulary of size $V$. For a selected set $S=\{p_1<\cdots<p_m\}\subseteq\{0,\ldots,N-1\}$, the receiver observes the selected token values and their zero-based positions. The packet uses one mode bit followed by either an $N$-bit bitmap or a gap list. With $b_m=\lceil\log_2(N+1)\rceil$ and $b_p=\lceil\log_2N\rceil$, the gap-list length is
\begin{equation}
\begin{aligned}
 B_{\rm gap}(S)&=b_m+\mathbf{1}_{m>0}\left(b_p+\sum_{j=2}^{m}\gamma(p_j-p_{j-1})\right),\\
 \gamma(u)&=2\lfloor\log_2u\rfloor+1.
\end{aligned}
 \label{eq:gap}
\end{equation}
where the cardinality field terminates the list: the empty set has no first-position field, and a nonempty list contains the first absolute position and exactly $m-1$ positive Elias-gamma-coded gaps. The position cost is therefore
\begin{equation}
 B_{\rm pos}(S)=1+\min\{N,B_{\rm gap}(S)\}.
 \label{eq:position}
\end{equation}
The one-bit mode selects the shorter representation, so bitmap and gap-list fields are never charged twice. The core packet includes a 32-bit header, the position description, a $\lceil\log_2 V\rceil$-bit codeword per selected token, and a 16-bit CRC. For the CIFAR-10/STL-10 systems with $V=32$, this payload is 5 bits per token. Forward-error protection is applied once to the complete core packet:
\begin{equation}
\begin{aligned}
 B_{\rm core}(S)&=32+B_{\rm pos}(S)+m\lceil\log_2V\rceil+16,\\
 B(S)=B_{\rm tx}(S)&=\left\lceil1.25B_{\rm core}(S)\right\rceil.
\end{aligned}
 \label{eq:bits}
\end{equation}
Equivalently, $B_{\rm fec}=B_{\rm tx}-B_{\rm core}$; the implementation uses this multiplicative form and rounds only after summing the core bits. The realized packet rate is
\begin{equation}
 R(S)=\frac{B(S)}{HW}\ \mathrm{bpp}.
 \label{eq:rate}
\end{equation}
In the CIFAR-10 implementation, $N=64$, $V=32$, and indices are zero-based. The encoder sorts the positions before coding, while the decoder reads the mode, then either all $N$ bitmap entries or $m$, the first position, and exactly $m-1$ gaps; empty, singleton, and full sets thus have unique decodable syntax.
For the coded-link experiment, the serialized information packet has $B_{\rm core}$ bits and the packet-accounting proxy is $B_{\rm tx}=\lceil1.25B_{\rm core}\rceil$. The physical LDPC implementation separately pads the information packet to
\[
 k=16\left\lceil B_{\rm core}/16\right\rceil,\qquad
 n=\operatorname{round}(k/0.8),
\]
so that the realized code rate is $R_c=k/n=0.8$. We therefore distinguish the selection rate $R_{\rm packet}=B_{\rm tx}/(HW)$ from the realized PHY rate $R_{\rm PHY}=n/(HW)$; padding and rate matching are not silently counted as a second FEC factor in Eq.~(\ref{eq:bits}). In the formal records, the $0.20$-bpp packets have $B_{\rm core}=159$--$163$, $B_{\rm tx}=199$--$204$, $k\in\{160,176\}$, and $n\in\{200,220\}$; the $0.32$-bpp packets have $B_{\rm core}=258$--$261$, $B_{\rm tx}=323$--$327$, $k=272$, and $n=340$. The resulting mean $(R_{\rm packet},R_{\rm PHY})$ values are $(0.1985,0.2146)$ for Local-MDL and $(0.1987,0.2147)$ for GCR-C at $0.20$ bpp, and $(0.3157,0.3320)$ for both methods at $0.32$ bpp.
For a budget cap $B_{\max}$, the feasible action set is
\begin{equation}
 \mathcal F(S,B_{\max})=\{i\notin S:B(S\cup\{i\})\leq B_{\max}\}.
 \label{eq:feasible}
\end{equation}
All selector comparisons use Eqs.~(\ref{eq:gap})--(\ref{eq:rate}) for packet-budget feasibility and report the realized packet rate $R_{\rm packet}$; the coded-link experiment additionally distinguishes the realized physical-layer rate $R_{\rm PHY}$.

\subsection{Channel abstraction and cognitive transmitter state}
The clean packet protocol above admits a generic receiver-available-set abstraction. For a channel realization $\omega$, let $\mathcal C_\omega(S)$ denote the set of selected token positions available to the receiver after packet transmission and decoding, and define
\begin{equation}
 S_{\rm rx}=\mathcal C_\omega(S),\qquad
 d_\omega(S)=d\!\left(\mathcal C_\omega(S)\right).
 \label{eq:channel-map}
\end{equation}
This mapping can represent position-level erasures or a whole-frame decoding failure. In the coded-link experiment, $\mathcal C_\omega(S)=S$ when CRC decoding succeeds and $\mathcal C_\omega(S)=\varnothing$ when it fails; the same definition also remains compatible with future burst-loss or token-level channel models. The receiver reconstructs from $S_{\rm rx}$ using the same masked prior.

For the coded-link experiment, the same serialized packet is passed through a coded bit-level channel. Bits are CRC-protected, mapped to QPSK, encoded with a 5G-NR LDPC block at an approximately $0.8$ code rate, and decoded with 20 belief-propagation iterations. We use complex AWGN and single-tap block-Rayleigh fading with perfect receiver CSI; a failed CRC therefore maps the selected set to $\varnothing$. This produces a reproducible PHY stress test while keeping the GCR-C selection frozen and channel independent.

The transmitter decision state is written as
\begin{equation}
 s_t=\bigl(S_t,B_{\rm rem,t},u_t,c_t\bigr),
 \label{eq:cognitive-state}
\end{equation}
where $S_t$ is the protected token set, $B_{\rm rem,t}$ is the remaining packet budget, $u_t$ collects local recoverability or receiver-state features, and $c_t$ is optional channel context such as an estimated erasure level. The four stages in Fig.~\ref{fig:cognitive-loop} are: (i) perception of $(S_t,B_{\rm rem,t},u_t,c_t)$; (ii) reasoning by proposal construction and $Q_B$ completion; (iii) action through the baseline-relative gate; and (iv) adaptation of the set, budget, and eligibility state. In the clean experiments, $c_t$ is a constant null context and no channel feedback is used. This is a deterministic decision loop, not a reinforcement-learning formulation.

\subsection{Receiver-side completion}
The masked prior $p_\theta$ receives known tokens and a mask at untransmitted positions. A deterministic one-pass completion produces $\hat z_{\bar S}$ and an image reconstruction
\begin{equation}
 \hat x(S)=D_\theta(z_S,\hat z_{\bar S}),
 \qquad d(S)=\operatorname{PSNR}(x,\hat x(S)).
 \label{eq:recon}
\end{equation}
The one-pass rule is held fixed in Local-MDL, GCR-C, and Random comparisons. It is intentionally simple so that the experiment measures selection policy rather than a changing decoder schedule.

\section{Gated Counterfactual Refinement}
\noindent
GCR-C operates as a correction layer on top of Local-MDL. At an eligible outer state, Local-MDL first provides the default action, and a compact proposal augments it with spatially complementary alternatives. Each proposal action is evaluated by fixing it as the next transmission and completing the remaining budget with Local-MDL. GCR-C replaces the default action only if the best candidate has positive full-budget advantage; after one accepted intervention, all remaining actions revert to Local-MDL.

\subsection{Local-MDL backbone}
At state $S$, the local recoverability score is the masked negative log-likelihood
\begin{equation}
 m_i(S)=-\log p_\theta(z_i\mid z_S),
 \qquad a_L(S)=\argmax_{i\in\mathcal F(S,B_{\max})}m_i(S).
 \label{eq:local}
\end{equation}
The sign follows a description-length interpretation: a token that is difficult to predict from the known set is a valuable token to protect. Local-MDL repeatedly applies Eq.~(\ref{eq:local}) until the packet budget is full. It is also the fallback when GCR-C is gated off.

\subsection{Baseline-relative fixed-budget value}
Let $\pi_L$ denote Local-MDL continuation. For a candidate action $a\in\mathcal F(S,B_{\max})$, define a horizon-$h$ value by applying $a$ once and then using Local-MDL for $h-1$ further actions:
\begin{equation}
 Q_h(a\mid S)=d\bigl(\pi_L^{h-1}(S\cup\{a\})\bigr),
 \label{eq:qh}
\end{equation}
where $Q_B$ completes until no further action is feasible. The baseline-relative advantage is
\begin{equation}
 A_h(a\mid S)=Q_h(a\mid S)-Q_h(a_L(S)\mid S).
 \label{eq:adv}
\end{equation}
This subtraction removes the large image- and rate-dependent variation in absolute PSNR. It also makes the decision operational: a candidate is useful only if it beats the action the system would already take. The evaluator is not a theorem about the globally optimal set; it is a matched-completion measurement under the declared receiver policy. One outer candidate evaluation may contain multiple Local continuation actions and multiple $p_\theta$ forward calls, so these events are counted separately rather than being collapsed into an unqualified ``call'' count.

For a channel-aware extension, use the receiver-available set mapping in Eq.~(\ref{eq:channel-map}) and define
\begin{equation}
 \begin{aligned}
 Q_B^{\rm ch}(a\mid S,c)
 &=\mathbb{E}_{\omega\sim p(\omega\mid c)}
 \left[d_\omega\!\left(\pi_L^B(S\cup\{a\})\right)\right],\\
 A_B^{\rm ch}(a\mid S,c)
 &=Q_B^{\rm ch}(a\mid S,c)-Q_B^{\rm ch}(a_L\mid S,c).
 \end{aligned}
 \label{eq:channel-q}
\end{equation}
Equation~(\ref{eq:channel-q}) makes the optional context $c_t$ operational: the evaluator can average the completed reconstruction over channel realizations rather than optimizing only the clean receiver. The present selector uses the clean $Q_B$ in Eq.~(\ref{eq:qh}); the coded-link experiment applies channel errors after selection as a frozen-policy validation. Thus $Q_B^{\rm ch}$ is a defined extension and not a claim that channel-aware selection has already been validated.

The candidate-level advantage can be negative, because an individually plausible action may be worse than the Local action. To separate this signed quantity from state-level opportunity, we define
\begin{equation}
 \begin{aligned}
 H_B(S)&=\max_{a\in\mathcal F(S,B_{\max})} A_B(a\mid S),\\
 \widetilde H_B(S)&=\max_{a\in\mathcal P(S)} A_B(a\mid S).
 \end{aligned}
 \label{eq:headroom}
\end{equation}
Because $a_L(S)\in\mathcal F(S,B_{\max})$ and $A_B(a_L\mid S)=0$, the exhaustive first-action reference headroom $H_B(S)$ is nonnegative even though individual candidate advantages and their averages need not be. This reference enumerates feasible first actions at the current state and applies the fixed Local-MDL continuation; it is not a global subset optimum. The gap $H_B(S)-\widetilde H_B(S)$ is the proposal regret used below.

\subsection{Multi-source proposal and gate}
Exhaustively evaluating all feasible actions is expensive. GCR-C therefore forms
\begin{equation}
 \mathcal P_C(S)=\operatorname{Top3}_{m}(S)\cup
 \operatorname{Top5}_{c}(S),
 \label{eq:proposal}
\end{equation}
where $m$ is the Local-MDL score and $c$ selects spatially complementary coverage candidates. Duplicate actions are removed. The Local action $a_L(S)$ is always inserted explicitly, so the proposal cannot discard the fallback. The proposal-source study also evaluates the P4 union, $\operatorname{Top3}_{m}\cup\operatorname{Top3}_{g}\cup\operatorname{Top2}_{c}$, as a source-ablation reference. The progress variable is
\[
 r(S_t)=\frac{|S_t|}{N},
\]
where $S_t$ is the selected set before the current outer action. An outer state is eligible for counterfactual gating iff $r(S_t)\leq\rho_{\max}$, where $\rho_{\max}=0.30$. Thus the eligible selected-count states are $0$--$19$ for the CIFAR-10 grid ($N=64$) and $0$--$43$ for the STL-10 grid ($N=144$). This parameter restricts when the gate may be evaluated; it is distinct from the one-intervention-per-image limit. The proposal and $Q_B$ are evaluated at eligible states for $0.20$ and $0.32$ bpp until either one intervention is accepted or the packet budget terminates; after the first accepted intervention, all remaining actions follow Local-MDL. The $0.44$ bpp point is not triggered.

The gate is
{\small
\begin{equation}
 a^*(S)=\begin{cases}
 \argmax_{a\in\mathcal P_C(S)} A_B(a\mid S),&\max_a A_B(a\mid S)>\delta,\\
 a_L(S),&\text{otherwise},
 \end{cases}
 \label{eq:gate}
\end{equation}
}
with $\delta=0$ and at most one intervention per image. Because $a_L$ is in $\mathcal P_C$, the computed reference policy is baseline-preserving under the same evaluator; the practical quality guarantee is limited to proposal and evaluator fidelity.

\noindent\textbf{Proposition 1 (baseline preservation under exact matched completion).}
Assume that $a_L(S)\in\mathcal P_C(S)$ and that $Q_B$ is the deterministic quality obtained by taking one action and then applying the same Local-MDL continuation until no feasible action remains. If GCR-C selects the proposal maximizer only when its value is strictly larger than $Q_B(a_L(S)\mid S)$, and otherwise selects $a_L(S)$, then its completed clean-channel quality is no smaller than Local-MDL under the same budget.

\emph{Proof.} Since $a_L(S)$ belongs to $\mathcal P_C(S)$, $\max_{a\in\mathcal P_C(S)}Q_B(a\mid S)\geq Q_B(a_L(S)\mid S)$. If the strict gate is not satisfied, GCR-C reproduces the baseline action and continuation. If it is satisfied, it selects a candidate with strictly larger matched-completion value. All pre-intervention actions coincide with the Local-MDL baseline; after the intervention, both candidate and baseline values are evaluated using the same Local-MDL continuation policy from their respective post-action states. Because at most one intervention is accepted, the terminal clean-channel quality is therefore no smaller than the matched Local-MDL baseline under the stated assumptions. The proposition is not a guarantee for an approximate horizon, learned value, stochastic decoder, mismatched channel evaluator, or global subset optimum.
The baseline-preservation property applies only to the terminal metric used by $Q_B$, which is PSNR in the present implementation; it does not imply non-degradation of external semantic or perceptual metrics.

GCR-C is a non-learned correction policy in the present study; learning a lower-cost approximation of the proposal/evaluator pipeline is left to future work.

\section{Experimental Setup and Evaluation Protocol}
\subsection{Data, model, and held-out evaluation}
For CIFAR-10 \cite{krizhevsky2009learning}, the system uses a 32-entry vector-quantized codebook over non-overlapping $4\times4$ RGB patches and a masked prior with 64 token positions. Thus $N=64$ is the number of positions, $V=32$ is the codebook size, and each transmitted token uses 5 payload bits. The codebook and prior are trained on 2,000 images, 500 images form the selector-development pool, and 100 images are reserved for validation. After the proposal, evaluator, gate, and operating-rate schedule are fixed on development/validation data, performance is reported on a disjoint 500-image held-out test split that is not used for policy selection.

The selected configuration uses $\mathcal P_C$ (Top-3 Local plus Top-5 coverage), $Q_B$, $\delta=0$, $\rho_{\max}=0.30$, at most one intervention, and eligible outer-state evaluation at $0.20$ and $0.32$ bpp; the gate is not evaluated at $0.44$ bpp. Eligibility is evaluated from $r(S_t)=|S_t|/N$ using the selected set before the current outer action, giving states $0$--$19$ for CIFAR-10 and $0$--$43$ for STL-10. At each eligible state, the gate evaluates the deduplicated proposal and stops after the first accepted intervention or budget termination. Split manifests and evaluation artifacts are released with the paper.

\subsection{Validation studies and baselines}
Before evaluation, we audited packet encode--decode consistency, candidate alignment, bit feasibility, deterministic Local-MDL actions, mask consistency, and explicit Local inclusion; full records are released. Headroom uses 50 images per rate with initial and early states clustered by image for bootstrap resampling. The horizon, proposal-source, and proposal-size studies retain the same packet accounting, continuation, decoder, and initial state, and use only development/validation data.

As a matched external baseline, we adapt the max--min diversity rule of DivPrune \cite{alvar2025divprune} to the same discrete-token and packet-feasible interface. Following the published rule, we compute cosine distances between pre-quantization raw $4\times4$ RGB patch vectors and select tokens by global max--min initialization followed by farthest-first max--min updates. Because packet cost depends on selected positions, each step is restricted to the exact feasible set $\mathcal F(S,B_{\max})$ in Eqs.~(\ref{eq:gap})--(\ref{eq:feasible}). This is a protocol-matched adaptation of the published diversity principle rather than a reproduction of the original LMM task pipeline: the tokenizer, codewords, receiver completion, packet syntax, and held-out image indices are unchanged, while the raw-patch feature mapping, distance, tie-breaking, and packet-feasibility adaptation are fixed without using held-out results. The P4 union is retained in the headroom analysis as the pre-selection proposal reference used to characterize proposal miss-rate before the source ablation; the final $\mathcal P_C$ allocation is selected subsequently using development/validation-only proposal-source comparisons.

\subsection{Metrics and computation accounting}
We report PSNR, SSIM, packet bpp (the realized $R_{\rm packet}$), selected-token count, interventions, and the following computation counters: $N_{\rm prop}$ is the number of deduplicated candidates formed for an image, $N_{\rm cf}$ is the number of candidate-level $Q_h/Q_B$ evaluations executed, $N_{\rm roll}$ is the number of Local actions appended inside those counterfactual continuations, $N_{\rm prior}$ is the number of actual $p_\theta$ forward calls, $N_{\rm dec}$ is the number of reconstruction forwards, and $N_{\rm int}$ counts accepted non-Local actions. $T_{\rm perf}$ is the mean synchronized time in the complete 500-image held-out evaluation, whereas $T_{\rm trace}$ is the median-of-three time in the fixed first-50 instrumentation subset; the latter includes repeated measurements and logging and is not a population estimate.

Validation timing uses 20 warm-up forwards, CUDA synchronization before and after timing, and includes proposal evaluation plus the final one-pass reconstruction; full traces are retained in the reproducibility package. The worst-case pre-intervention candidate-evaluation bound follows directly from the eligibility rule and the maximum eight-action deduplicated proposal:
\[
  N_{\rm cf}^{\max}=8\left(\left\lfloor0.30N\right\rfloor+1\right),
\]
which gives 160 evaluations for CIFAR-10 ($N=64$) and 352 for STL-10 ($N=144$). The realized number is lower when the packet budget terminates early, proposal deduplication reduces the candidate count, or an intervention is accepted before all eligible states are visited. For statistical analysis, paired comparisons use image-level bootstrap resampling; headroom states from the same image are retained together. The reported interval is the percentile 95\% CI.

\subsection{Link-level and high-resolution transfer protocols}
The link-level experiment reuses the frozen clean selections from the 500-image CIFAR-10 evaluation; no $Q_B$ decision is re-optimized for a particular SNR. We implement the packet link with Sionna 2.0.1 \cite{hoydis2022sionna}: CRC-protected packets are encoded by a 5G-NR LDPC block with a code rate close to $0.8$, mapped to QPSK, and decoded by 20 belief-propagation iterations. For each coded block, the complex AWGN variance is set from the information-bit $E_b/N_0$ as
\[
 \sigma^2=\left[R_c\log_2(M)10^{(E_b/N_0)/10}\right]^{-1},
 \qquad M=4,
\]
and the equalized block-Rayleigh branch uses $\sigma^2/|h|^2$ with perfect receiver CSI. We test complex AWGN and single-tap block-Rayleigh fading. A 50-image validation pilot sweep uses a $0.5$-dB grid; for each rate and channel, hard/medium/easy points are selected from distinct observed SNRs by nearest FER to $0.525/0.30/0.075$ in hard-to-easy order. The selected points are then frozen and evaluated on all 500 held-out images. Each image--condition pair uses one channel realization ($R=1$); the deterministic fading/noise seed is shared between Local-MDL and GCR-C, and paired inference is performed before the image-level bootstrap. A CRC failure is counted as a frame error and exposes no selected token to the receiver, which then uses the same masked prior. Thus the experiment evaluates a frozen selection over a coded PHY link, rather than a channel-aware selector or an over-the-air system.

For a higher-resolution transfer check, we use the official LlamaGen VQ-16 tokenizer \cite{llamagen2024} with a 16-pixel downsampling ratio, codebook size $V=16{,}384$, embedding dimension 8, and a $24\times24$ grid ($N=576$) for $384\times384$ center-cropped Kodak-24 images \cite{kodak24}. A lightweight six-layer, six-head masked prior of width 384 is trained from scratch on four deterministic crops of each of 800 DIV2K training images (3,200 sequences); 20 separate DIV2K images are used for validation. Two packet budgets are calibrated from eight validation images using the 75th percentile of exact packet bits, giving 2,101 bits (HR-Low) and 3,804 bits (HR-Mid). Because a complete outer-state $Q_B$ sweep is substantially more expensive at $N=576$, this transfer diagnostic uses a pre-registered initial-state Top-3 Local plus Top-5 coverage gate, at most one intervention, and Local continuation thereafter; the CIFAR-10 and STL-10 core protocols retain the full eligible-state rule in Sec.~IV. This experiment is used as a transfer diagnostic for the reconstruction-oriented decision mechanism under a substantially larger token grid and vocabulary. Separate semantic/perceptual diagnostics are provided in the supplementary material and are not used for candidate selection or policy calibration.

\section{Results}
\subsection{Rate-dependent counterfactual headroom}
Counterfactual headroom is concentrated at low rate but remains nonzero at the highest tested rate. At $0.20$ bpp, the mean exhaustive first-action headroom is 2.45--2.58 dB and 84--86\% of states exceed 0.10 dB. At $0.32$ bpp, the mean falls to 0.73--0.81 dB and 52\% exceed 0.10 dB; at $0.44$ bpp, the mean is 0.27 dB and 34\% exceed 0.10 dB. The intervals for the mean $H_B$ are image-cluster bootstrap intervals, not state-level independent intervals. Although $Q_B$ remains positive at $0.44$ bpp (+0.091 dB, 95\% CI $[0.023,0.183]$), the synchronized validation cost is approximately 2.61 s/image. The validation-selected operating schedule therefore disables counterfactual evaluation at this rate. The separation in Eq.~(\ref{eq:headroom}) remains important: first-action reference headroom is nonnegative by construction, whereas the candidate distribution can have a negative mean.

Complete state-level headroom and P4-regret statistics are provided in Supplementary Table~S3; Fig.~\ref{fig:headroom} retains the distributional and mechanism-level evidence in the main paper.

The proposal analysis also shows that the compact candidate set does not enumerate all feasible first actions: P4 regret is 0.95 dB at $0.20$ bpp, 0.45--0.57 dB at $0.32$ bpp, and 0.18--0.22 dB at $0.44$ bpp in these states. Coverage candidates recover gaps left by Local and set-gain sources, but the remaining regret bounds the interpretation. Accordingly, we interpret GCR-C as a gated correction layer rather than an optimal subset selector.

\begin{figure*}[tbp]
\centering
\includegraphics[width=0.88\textwidth]{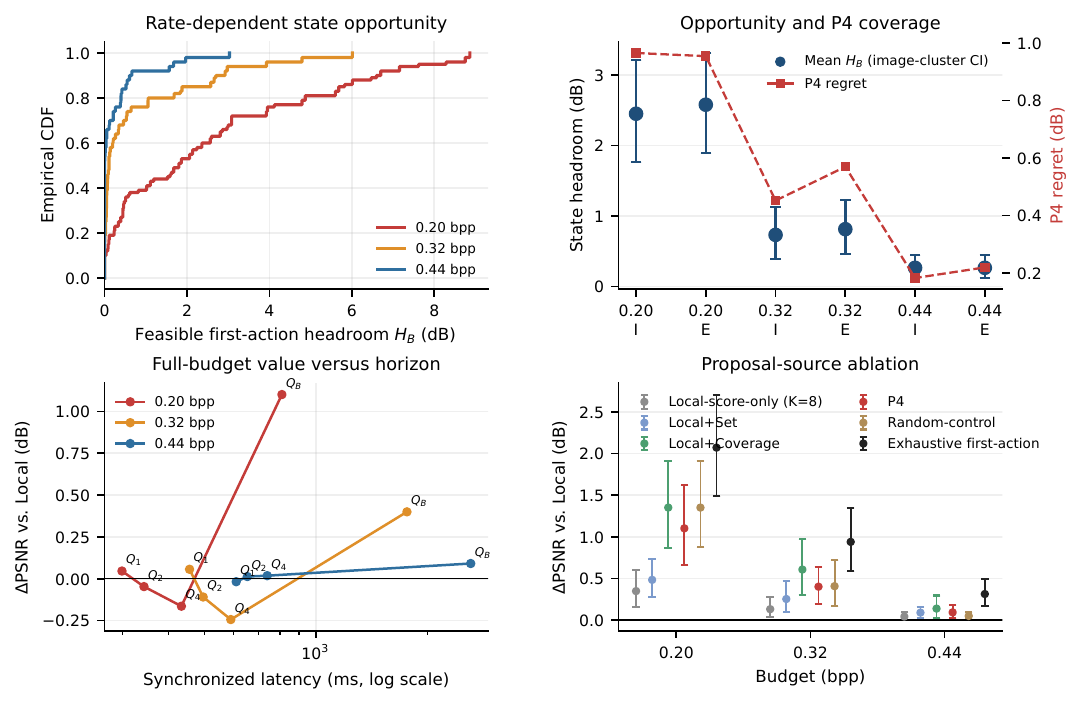}
\caption{Validation diagnostics: headroom CDF (a), state headroom/P4 regret with image-cluster CIs (b), horizon quality--latency frontier (c), and proposal-source ablation (d).}
\label{fig:headroom}
\end{figure*}

\subsection{Full-budget value versus short-horizon approximations}
Short-horizon value estimates do not reliably recover the full-budget decision, despite their lower latency. $Q_1$, $Q_2$, and $Q_4$ recover at most 14.1\% of the $Q_B$ gain at $0.32$ bpp and can be worse than Local after the final-budget continuation; their confidence intervals include zero at the active rates. The quality-first $Q_B$ mode is the only horizon with a positive interval at both active rates, although it is not a low-latency approximation: its validation recovery is 1.0 by definition and its synchronized latency is 0.81--1.76 s per image at $0.20$--$0.32$ bpp. At $0.44$ bpp, $Q_B$ still gives +0.091 dB (95\% CI $[0.023,0.183]$) but costs about 2.61 s/image. We therefore use $Q_B$ as the reference-quality evaluator and treat shorter horizons as lower-cost alternatives rather than substitutes for the full-budget decision.

The synchronized validation counter measurements quantify the cost of the quality-first evaluator. In the initial-state diagnostic, each formed proposal candidate is evaluated exactly once, so $N_{\rm prop}=N_{\rm cf}=7.6$ on average. The full-budget evaluator adds 93.1, 221.3, and 372.4 counterfactual Local rollout actions at the three rates. These measurements make the encoder-side computation visible alongside the quality evidence; the multi-state operating rule can accumulate more than one proposal size before the first accepted intervention.

\subsection{Proposal-source ablation}
Diversified proposals improve access to counterfactual headroom, but the data do not support a universally superior coverage heuristic. Local+Coverage is the strongest compact source variant at the two active rates, improving over the previous P4 union by 0.250 dB at $0.20$ bpp (paired 95\% CI $[0.025,0.513]$) and 0.206 dB at $0.32$ bpp (CI $[0.020,0.477]$). However, the direct paired Local+Coverage versus Random-control difference is only $+0.000$ dB (CI $[-0.333,0.321]$) at $0.20$ bpp and $+0.200$ dB (CI $[-0.019,0.504]$) at $0.32$ bpp; neither interval is strictly positive. We therefore interpret the result as evidence for diversified proposals, with coverage-aware diversification becoming more advantageous at the medium rate, rather than as proof of a coverage-specific universal mechanism. The exhaustive row is an offline first-action reference under the Local-MDL continuation: it enumerates feasible actions at the current intervention state but does not solve the global budgeted subset-selection problem. It uses up to 64 feasible first actions and costs 4.38, 10.77, and 16.32 s per image at the three rates.

\subsection{Proposal-size quality--compute frontier}
$K=8$ lies near the main knee of the proposal-size quality--compute frontier. $K=12$ improves over $K=8$ by 0.202 dB at $0.20$ bpp, but only 0.108 and 0.049 dB at $0.32$ and $0.44$ bpp while increasing synchronized latency by approximately 27--33\%; $K=16$ adds at most 0.044, 0.009, and 0.012 dB beyond $K=12$. The compact $K=8$ point is therefore used for the final proposal. Within that budget, the source study selects a coverage-aware diversification allocation, yielding $\mathcal P_C$ rather than the previous P4 union.

\begin{figure*}[tbp]
\centering
\includegraphics[width=0.88\textwidth]{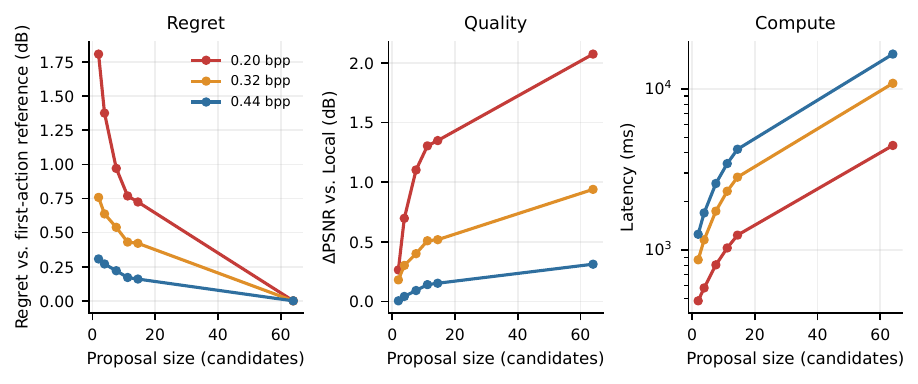}
\caption{Proposal-size quality--compute frontier; $K=8$ is the final compact operating point.}
\label{fig:size}
\end{figure*}

\begin{table*}[tbp]
\centering
\caption{Held-out CIFAR-10 results on 500 images. Values are mean $\pm$ standard deviation for PSNR and mean SSIM. Random, Entropy-greedy, Coverage-greedy, and Adapted max--min diversity are matched baselines evaluated with the same packet protocol; the difference, CI, and win rate are paired against Local-MDL. At 0.44 bpp, GCR-C is identical to Local-MDL by policy; therefore, the win rate is 0 and the tie rate is 1.00.}
\label{tab:final}
\scriptsize
\resizebox{\textwidth}{!}{%
\begin{tabular}{c|l|c|c|c|c|c|c|c}
\toprule
Nominal bpp & Method & Packet bpp & Tokens & PSNR (dB) & SSIM & $\Delta$ PSNR & 95\% CI & Win rate \\
\midrule
0.20 & Random & 0.1986 & 11.41 & 13.215$\pm$3.511 & .382 & +0.798 & [+.552,+1.042] & .548 \\
      & Entropy-greedy & 0.1986 & 12.37 & 13.319$\pm$3.520 & .392 & +0.902 & [+.674,+1.134] & .576 \\
      & Coverage-greedy & 0.1992 & 11.00 & 12.889$\pm$3.653 & .365 & +0.472 & [+.237,+.706] & .506 \\
      & Adapted max--min diversity \cite{alvar2025divprune} & 0.1984 & 12.49 & 11.903$\pm$4.006 & .333 & $-$0.514 & [$-$0.775,$-$0.245] & .368 \\
      & Local-MDL & 0.1985 & 12.02 & 12.417$\pm$3.864 & .395 & 0 & --- & --- \\
      & GCR-C & 0.1987 & 12.01 & 13.920$\pm$3.637 & .436 & +1.503 & [+1.341,+1.670] & .974 \\
\midrule
0.32 & Random & 0.3154 & 29.00 & 14.629$\pm$3.232 & .479 & $-$0.266 & [$-$0.484,$-$0.046] & .398 \\
      & Entropy-greedy & 0.3163 & 29.41 & 14.919$\pm$3.484 & .513 & +0.024 & [$-$0.177,+.224] & .448 \\
      & Coverage-greedy & 0.3154 & 29.00 & 14.147$\pm$3.552 & .454 & $-$0.748 & [$-$0.979,$-$0.514] & .352 \\
      & Adapted max--min diversity \cite{alvar2025divprune} & 0.3159 & 29.21 & 13.909$\pm$4.205 & .464 & $-$0.986 & [$-$1.215,$-$0.757] & .306 \\
      & Local-MDL & 0.3157 & 29.11 & 14.895$\pm$3.892 & .529 & 0 & --- & --- \\
      & GCR-C & 0.3157 & 29.11 & 15.593$\pm$3.605 & .550 & +0.698 & [+.595,+.809] & .904 \\
\midrule
0.44 & Random & 0.4375 & 49.00 & 16.873$\pm$2.761 & .606 & $-$1.716 & [$-$1.893,$-$1.534] & .146 \\
      & Entropy-greedy & 0.4375 & 49.00 & 17.471$\pm$2.890 & .646 & $-$1.118 & [$-$1.282,$-$0.949] & .152 \\
      & Coverage-greedy & 0.4375 & 49.00 & 16.456$\pm$3.277 & .606 & $-$2.133 & [$-$2.368,$-$1.903] & .164 \\
      & Adapted max--min diversity \cite{alvar2025divprune} & 0.4375 & 49.00 & 16.876$\pm$3.563 & .623 & $-$1.713 & [$-$1.918,$-$1.511] & .134 \\
      & Local-MDL & 0.4375 & 49.00 & 18.589$\pm$2.902 & .680 & 0 & --- & --- \\
      & GCR-C & 0.4375 & 49.00 & 18.589$\pm$2.902 & .680 & 0 & [0,0] & 0 \\
\bottomrule
\end{tabular}}
\end{table*}

\subsection{Held-out CIFAR-10 performance}
GCR-C converts validation-observed low-rate headroom into held-out PSNR gains without increasing realized packet rate. It improves over Local-MDL by 1.503 dB at $0.20$ bpp and 0.698 dB at $0.32$ bpp, with strictly positive paired intervals and win rates of 97.4\% and 90.4\%, respectively. At $0.44$ bpp, the selected operating rule does not evaluate the gate, so GCR-C follows Local-MDL and the two methods are identical. The fixed heuristics are informative but do not reproduce the same pattern: Entropy-greedy gives +0.902 dB at $0.20$ bpp but only +0.024 dB with a CI crossing zero at $0.32$ bpp, while Coverage-greedy gives +0.472 and $-$0.748 dB. Thus the active-rate result is not explained by a static entropy or coverage score alone.

The adapted max--min diversity baseline achieves 11.903, 13.909, and 16.876 dB PSNR at $0.20$, $0.32$, and $0.44$ bpp, respectively, corresponding to $-$0.514, $-$0.986, and $-$1.713 dB relative to Local-MDL. At the two active rates, the direct paired differences between GCR-C and the adapted max--min diversity baseline are +2.017 dB (95\% CI $[+1.777,+2.258]$) and +1.684 dB (CI $[+1.459,+1.904]$). The adapted max--min diversity selector therefore does not reproduce the active-rate baseline-relative full-budget correction, supporting the counterfactual value comparison rather than a generic diversity-only explanation.

\begin{figure*}[tbp]
\centering
\includegraphics[width=0.88\textwidth]{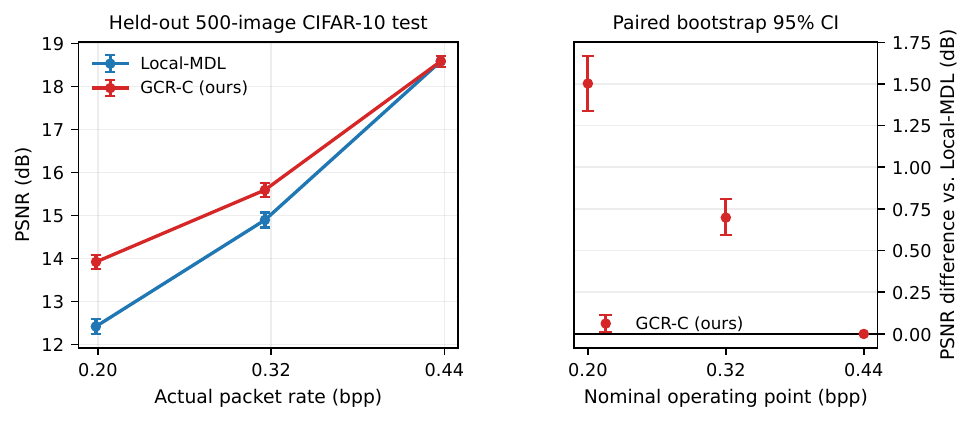}
\caption{Held-out CIFAR-10 rate--distortion and paired GCR-C--Local-MDL 95\% CIs; matched baselines are in Table~\ref{tab:final}.}
\label{fig:final}
\end{figure*}

\subsection{Frozen selections over a coded wireless link}
The coded-link experiment tests whether the clean-selection advantage survives packet decoding errors without changing either selector. Across two rates, two channels, and three distinct validation-pilot-calibrated $E_b/N_0$ points per channel, GCR-C has a positive paired PSNR gain at all 12 conditions, ranging from $+0.378$ to $+1.416$ dB; every image-level 95\% bootstrap interval remains above zero. Since packet lengths are nearly identical, the gain reflects the reconstruction obtained from tokens that pass decoding rather than a larger transmitted payload. At $0.20$ bpp, the AWGN gains are $+0.946$--$+1.416$ dB and the block-Rayleigh gains are $+0.697$--$+1.304$ dB; at $0.32$ bpp, the corresponding ranges are $+0.378$--$+0.647$ and $+0.394$--$+0.630$ dB. This is a frozen-policy coded-link validation, not a channel-aware selector or an over-the-air measurement.

\begin{table}[tbp]
\centering
\caption{Compact frozen-selection coded-link summary on the held-out CIFAR-10 split. $E_b/N_0$ and $\Delta$PSNR are listed as hard/medium/easy triples; all 12 paired 95\% CIs are positive. Complete FER, PSNR, and CI records are provided in Supplementary Table~S4.}
\label{tab:wireless-phy}
\scriptsize
\setlength{\tabcolsep}{2.5pt}
\renewcommand{\arraystretch}{0.92}
\begin{tabular}{@{}llcc@{}}
\toprule
Rate & Channel & $E_b/N_0$ (H/M/E) & $\Delta$PSNR (H/M/E) \\
\midrule
0.20 & AWGN & 3.0/3.5/4.5 & +.946/+1.194/+1.416 \\
0.20 & Block-Rayleigh & 4.0/8.0/12.5 & +.697/+1.071/+1.304 \\
0.32 & AWGN & 2.5/3.0/3.5 & +.378/+.539/+.647 \\
0.32 & Block-Rayleigh & 4.0/7.5/12.0 & +.394/+.450/+.630 \\
\bottomrule
\end{tabular}
\end{table}

\subsection{Realized rate and encoder-side cost}
The quality gain is accompanied by substantial encoder-side full-budget evaluation rather than additional transmitted payload. In the held-out evaluation, GCR-C executes 13.198 and 26.876 counterfactual candidate evaluations per image at $0.20$ and $0.32$ bpp, respectively, compared with zero for Local-MDL and Random. These are $N_{\rm cf}$, not $N_{\rm prior}$ or Local rollout actions. The intervention rates are 0.974 and 0.904, and GCR-C selects 12.01 and 29.11 tokens on average, so the gains are not caused by a larger nominal payload. The synchronized measurements give $T_{\rm perf}$ means of 883 ms and 3945 ms at the two active rates.

The fixed first-50 instrumentation subset gives $N_{\rm prop}=N_{\rm cf}=18.98/58.60$, $N_{\rm roll}=154.34/1174.60$, $N_{\rm prior}=41.68/202.80$, and $N_{\rm dec}=5.88/15.84$ at $0.20/0.32$ bpp; the mean per-image median-of-three synchronized times, denoted $T_{\rm trace}$, are 1565/10403 ms. These diagnostic values are not population estimates and are not averaged with the held-out means. The RTX 3060 measurements quantify encoder-side cost for the unbatched evaluator rather than hardware-independent latency.

All rows use Eq.~(\ref{eq:bits}), including packet, position, payload, CRC, and FEC terms. The close packet-bpp values for Local-MDL and GCR-C show that the gains are not caused by sending a larger nominal payload. At $0.20$ bpp, GCR-C selects 12.01 tokens on average versus 12.02 for Local-MDL; at $0.32$ bpp the corresponding values are 29.11 and 29.11.

Together, the results indicate that GCR-C is most useful when substantial full-budget headroom remains and a compact proposal can expose alternatives to the Local action. Its gain therefore depends jointly on rate, proposal coverage, and evaluator fidelity. The horizon study further shows that short-horizon scores are lower-cost approximations but do not reproduce the full-budget decision in the present system.

\subsection{Cross-dataset replication on STL-10}
The correction mechanism persists after changing dataset, spatial resolution, token grid, and independently trained prior. STL-10 \cite{coates2011analysis} uses native $96\times96$ RGB images, $8\times8$ patches, $N=144$ positions, and a newly trained $V=32$ codebook/prior. Five thousand unlabeled images train the visual prior, 500 additional unlabeled images are reserved for prior validation, and 200 labeled training images calibrate the two bit budgets. A stratified 500-image subset of the official test set is used for held-out evaluation. The same decision design is applied with budget caps of 350 and 650 bits; eligibility remains $|S_t|/N\leq0.30$, corresponding to selected-count states $0$--$43$ for $N=144$. The tokenizer and prior are independently trained, so this is cross-dataset replication of the decision mechanism rather than zero-shot checkpoint transfer.

The paired gain is $+0.621$ dB at the lower budget (95\% CI $[0.559,0.687]$, win rate 99.6\%) and $+0.204$ dB at the middle budget (CI $[0.169,0.243]$, win rate 94.2\%). These intervals come from 10,000 image-level paired bootstrap resamples, and the held-out images are not used for checkpoint, budget, or policy selection. The result provides limited external-validity evidence that the baseline-relative correction persists under a changed resolution, token grid, and independently trained prior.

\begin{table}[tbp]
\centering
\caption{STL-10 replication on a stratified 500-image held-out subset. Packet bpp is recomputed from complete packet fields; $\Delta$PSNR, CI, and win rate are paired GCR-C--Local-MDL differences.}
\label{tab:stl10}
\scriptsize
\setlength{\tabcolsep}{2pt}
\renewcommand{\arraystretch}{0.8}
\resizebox{\columnwidth}{!}{%
\begin{tabular}{@{}llccccccc@{}}
\toprule
Point & Method & \shortstack{Cap\\(bits)} & \shortstack{Packet\\bpp} & Tokens & \shortstack{PSNR/\\SSIM} & $\Delta$PSNR & 95\% CI & Win \\
\midrule
Low & Local-MDL & 350 & .037907 & 24.912 & 11.998/.261 & -- & -- & -- \\
Low & GCR-C & 350 & .037911 & 24.754 & 12.620/.275 & +.621 & [.559,.687] & 99.6\% \\
Mid & Local-MDL & 650 & .070346 & 65.678 & 14.533/.358 & -- & -- & -- \\
Mid & GCR-C & 650 & .070350 & 65.630 & 14.738/.364 & +.204 & [.169,.243] & 94.2\% \\
\bottomrule
\end{tabular}
}
\end{table}

\subsection{High-resolution transfer on Kodak-24}
The Kodak experiment serves as a high-resolution transfer diagnostic rather than a semantic-codec benchmark. With the token grid enlarged to $24\times24$ ($N=576$) and the vocabulary to $V=16{,}384$, GCR-C changes the realized packet rate by less than $3\times10^{-6}$ bpp relative to Local-MDL while improving mean PSNR by $+1.033$ dB at HR-Low and $+0.847$ dB at HR-Mid. The paired 95\% confidence intervals remain above zero, with image-level win rates of $83.3\%$ and $70.8\%$, respectively. These results provide protocol-scoped evidence that the baseline-relative correction mechanism transfers to a substantially larger discrete visual representation; they are not evidence that the lightweight high-resolution receiver is a competitive image codec. Separate DINOv2/LPIPS diagnostics are reported in the supplementary material and do not establish a consistent semantic or perceptual advantage; accordingly, no such advantage is claimed here.

\begin{table}[tbp]
\centering
\caption{Kodak-24 high-resolution transfer. Values are means over 24 images; Packet bpp is realized packet rate and the $\Delta$PSNR interval is a paired 10,000-resample image-level bootstrap CI.}
\label{tab:highres-transfer}
\scriptsize
\setlength{\tabcolsep}{2.5pt}
\renewcommand{\arraystretch}{0.92}
\resizebox{\columnwidth}{!}{%
\begin{tabular}{@{}llrrcc@{}}
\toprule
Point & Method & Tok. & Pkt bpp & PSNR/SSIM & $\Delta$PSNR [CI] \\
\midrule
HR-Low & Local-MDL & 87.708 & .014204 & 9.349/.241 & -- \\
        & GCR-C     & 87.667 & .014201 & 10.382/.273 & +1.033 [.641,1.456] \\
\midrule
HR-Mid & Local-MDL & 173.500 & .025748 & 10.329/.238 & -- \\
        & GCR-C     & 173.542 & .025746 & 11.177/.290 & +.847 [.369,1.417] \\
\bottomrule
\end{tabular}
}
\end{table}

\section{Limitations and Reproducibility}
\subsection{Limitations}
The evidence has five boundaries. First, CIFAR-10/STL-10 use compact tokenizers; the Kodak prior is lightweight and its weak low-rate reconstructions do not establish transfer to full LlamaGen/VQGAN/diffusion/multimodal tokenizers. Second, Kodak uses an initial-state gate with Local continuation, so it is a transfer diagnostic rather than the full eligible-state schedule. Third, the unbatched full-budget evaluator is expensive and is best viewed as a quality-first operating point for delay-tolerant systems with encoder compute. Fourth, the coded link is complex AWGN/single-tap block-Rayleigh with perfect CSI and excludes mobility, burst errors, feedback/multi-user effects, and OTA impairments; matched receiver-side tokenizer/prior/decoder access and model-mismatch effects are not studied. Fifth, $Q_B$ is PSNR-specific: Kodak DINOv2/LPIPS do not show consistent advantage, so pixel-domain baseline preservation is not semantic-metric preservation. Task-aware terminal utilities, stronger priors, learned/batched evaluation, OTA validation, and channel-aware $Q_B^{\rm ch}$ selection remain future work. The adapted max--min result is a protocol-matched adaptation, not a reproduction of the original LMM pipeline.

\subsection{Reproducibility and Data Availability}
The package contains checkpoints, split manifests, fixed policies, scripts, result records, coded-link/Kodak calibration files, figures, computation traces, and paired-bootstrap summaries. Policies were frozen on development/validation data before held-out evaluation; coded-link trials use shared seeds. CIFAR-10, STL-10, and Kodak-24 follow the public sources \cite{krizhevsky2009learning,coates2011analysis,kodak24}; full file-level metric and protocol paths, including the complete headroom and coded-link audits, are provided in Supplementary Tables~S3--S4 and the package.

\section{Conclusion}
We introduced GCR-C, a communication-specific rollout-style correction layer that evaluates feasible alternatives with same-budget Local continuation and accepts only positive full-budget advantage. On held-out CIFAR-10 it improves PSNR by 1.503 and 0.698 dB at 0.20/0.32 bpp; the 0.44-bpp policy follows Local-MDL. Under frozen coded-link selections, paired gains are positive at all 12 AWGN/block-Rayleigh conditions. Independent STL-10 and limited Kodak transfer provide protocol-scoped evidence across dataset, resolution, token grid, and tokenizer. The remaining cost is encoder-side full-budget evaluation and no OTA validation; batching, learned value approximations, stronger priors, and channel-aware selection are next steps. Semantic/perceptual improvement is not claimed because it is not the present terminal utility.

\section*{AI Use Disclosure}
Artificial intelligence (AI) tools were used solely for language editing and polishing.


\begin{thebibliography}{43}
\scriptsize
\bibitem{edwards2024deepspace}
B. L. Edwards, D. Antsos, A. Biswas, R. Reinhart, B. Robinson,
D. Boroson, F. Khatri, and S. Lichten,
``Addressing the high-rate deep space communications shortfall in
NASA's Space Technology Mission Directorate's envisioned future,''
in \emph{Proc. 29th Ka and Broadband Communications Conference},
Seattle, WA, USA, 2024.

\bibitem{lin2011spacecube}
M. Lin, T. Flatley, J. Godfrey, A. Geist, D. Espinosa, and D. Petrick,
``SpaceCube 2.0: An advanced hybrid onboard data processor,''
\emph{NASA Tech Briefs}, Feb. 2011, NASA Technical Reports Server
Document ID 20110012222.

\bibitem{shannon1948mathematical} C. E. Shannon, ``A mathematical theory of communication,'' \emph{Bell Syst. Tech. J.}, vol. 27, no. 3, pp. 379--423, 1948.
\bibitem{oord2017neural} A. van den Oord, O. Vinyals, and K. Kavukcuoglu, ``Neural discrete representation learning,'' arXiv:1711.00937, 2017.
\bibitem{esser2021taming} P. Esser, R. Rombach, and B. Ommer, ``Taming transformers for high-resolution image synthesis,'' in \emph{Proc. IEEE/CVF CVPR}, 2021.
\bibitem{chang2022maskgit} H. Chang, H. Zhang, L. Jiang, C. Liu, and W. T. Freeman, ``MaskGIT: Masked generative image transformer,'' in \emph{Proc. IEEE/CVF CVPR}, pp. 11315--11325, 2022.
\bibitem{bourtsoulatze2019deep} E. Bourtsoulatze, D. B. Kurka, and D. Gunduz, ``Deep joint source-channel coding for wireless image transmission,'' \emph{IEEE Trans. Cogn. Commun. Netw.}, vol. 5, no. 3, pp. 567--579, 2019.
\bibitem{xie2021deep} H. Xie, Z. Qin, G. Y. Li, and B.-H. Juang, ``Deep learning enabled semantic communication systems,'' \emph{IEEE Trans. Signal Process.}, vol. 69, pp. 2663--2675, 2021.
\bibitem{han2022generative} T. Han, J. Tang, Q. Yang, Y. Duan, Z. Zhang, and Z. Shi, ``Generative model based highly efficient semantic communication approach for image transmission,'' arXiv:2211.10287, 2022.
\bibitem{pezone2025sqgan} F. Pezone, S. Barbarossa, and G. Caire, ``SQ-GAN: Semantic image communications using masked vector quantization,'' \emph{IEEE Trans. Cogn. Commun. Netw.}, early access, 2025, doi: 10.1109/TCCN.2025.3620819.
\bibitem{bertsekas2019biased} D. P. Bertsekas, ``Biased aggregation, rollout, and enhanced policy improvement for reinforcement learning,'' arXiv:1910.02426, 2019.
\bibitem{wang2026wmcdt} L. Wang, T. Shui, W. Saad, and P. Adjakple, ``World model-enabled causal digital twins for semantic communications in physical AI systems,'' arXiv:2605.16547, 2026.
\bibitem{rao2021dynamicvit} Y. Rao, W. Zhao, B. Liu, J. Lu, J. Zhou, and C.-J. Hsieh, ``DynamicViT: Efficient vision transformers with dynamic token sparsification,'' in \emph{Advances in Neural Information Processing Systems}, vol. 34, 2021.
\bibitem{krizhevsky2009learning} A. Krizhevsky, ``Learning multiple layers of features from tiny images,'' Univ. Toronto, Tech. Rep., 2009.
\bibitem{coates2011analysis} A. Coates, H. Lee, and A. Y. Ng, ``An analysis of single-layer networks in unsupervised feature learning,'' in \emph{Proc. AISTATS}, vol. 15, pp. 215--223, 2011.
\bibitem{rissanen1989stochastic} J. Rissanen, \emph{Stochastic Complexity in Statistical Inquiry}. Singapore: World Scientific, 1989.
\bibitem{minnen2018joint} D. Minnen, J. Ball{\'e}, and G. D. Toderici, ``Joint autoregressive and hierarchical priors for learned image compression,'' in \emph{Advances in Neural Information Processing Systems}, vol. 31, 2018.
\bibitem{li2025inference} K. Y. Li, S. Goyal, J. D. Semedo, and J. Z. Kolter, ``Inference optimal VLMs need fewer visual tokens and more parameters,'' in \emph{Proc. ICLR}, 2025.
\bibitem{ye2025voco} X. Ye, Y. Gan, X. Huang, Y. Ge, and Y. Tang, ``VoCo-LLaMA: Towards vision compression with large language models,'' in \emph{Proc. IEEE/CVF CVPR}, pp. 29836--29846, 2025.
\bibitem{yang2025pvc} C. Yang, X. Dong, X. Zhu, W. Su, J. Wang, H. Tian, Z. Chen, W. Wang, L. Lu, and J. Dai, ``PVC: Progressive visual token compression for unified image and video processing in large vision-language models,'' in \emph{Proc. IEEE/CVF CVPR}, pp. 24939--24949, 2025.
\bibitem{alvar2025divprune} S. R. Alvar, G. Singh, M. Akbari, and Y. Zhang, ``DivPrune: Diversity-based visual token pruning for large multimodal models,'' in \emph{Proc. IEEE/CVF CVPR}, pp. 9392--9401, 2025.
\bibitem{bergner2025tokencropr} B. Bergner, C. Lippert, and A. Mahendran, ``Token Cropr: Faster ViTs for quite a few tasks,'' in \emph{Proc. IEEE/CVF CVPR}, pp. 9740--9750, 2025.
\bibitem{dhouib2025pact} M. Dhouib, D. Buscaldi, S. Vanier, and A. Shabou, ``PACT: Pruning and clustering-based token reduction for faster visual language models,'' in \emph{Proc. IEEE/CVF CVPR}, pp. 14582--14592, 2025.
\bibitem{choi2025representationshift} J. Choi, S. Lee, B. Ko, E. Kim, J. Kil, and H. J. Kim, ``Representation shift: Unifying token compression with FlashAttention,'' in \emph{Proc. IEEE/CVF ICCV}, pp. 20456--20466, 2025.
\bibitem{kim2025tokenredundancy} K. Kim, J. Park, J. Kim, H. Kwon, and K. Sohn, ``Faster parameter-efficient tuning with token redundancy reduction,'' in \emph{Proc. IEEE/CVF CVPR}, pp. 30189--30198, 2025.
\bibitem{zeng2025skipvision} W. Zeng, Z. Huang, K. Ji, and Y. Yan, ``Skip-Vision: Efficient and scalable acceleration of vision-language models via adaptive token skipping,'' in \emph{Proc. IEEE/CVF ICCV}, pp. 21384--21397, 2025.
\bibitem{ye2025atp} X. Ye, Y. Gan, Y. Ge, X.-P. Zhang, and Y. Tang, ``ATP-LLaVA: Adaptive token pruning for large vision language models,'' in \emph{Proc. IEEE/CVF CVPR}, pp. 24972--24982, 2025.
\bibitem{endo2025feather} M. Endo, X. Wang, and S. Yeung-Levy, ``Feather the throttle: Revisiting visual token pruning for vision-language model acceleration,'' in \emph{Proc. IEEE/CVF ICCV}, pp. 22826--22835, 2025.
\bibitem{liu2025kvtp} Y. Liu, J. Sun, Y. Lin, J. Zhang, J. Zhang, M. Yin, Q. Wang, H. Li, and Y. Chen, ``Keyframe-oriented vision token pruning: Enhancing efficiency of large vision-language models on long-form video processing,'' in \emph{Proc. IEEE/CVF ICCV}, pp. 20802--20811, 2025.
\bibitem{wang2025dynamicvlm} H. Wang, Y. Nie, Y. Ye, Y. Wang, S. Li, H. Yu, J. Lu, and C. Huang, ``Dynamic-VLM: Simple dynamic visual token compression for VideoLLM,'' in \emph{Proc. IEEE/CVF ICCV}, pp. 20812--20823, 2025.
\bibitem{zha2025textok} K. Zha, L. Yu, A. Fathi, D. A. Ross, C. Schmid, D. Katabi, and X. Gu, ``Language-guided image tokenization for generation,'' in \emph{Proc. IEEE/CVF CVPR}, pp. 15713--15722, 2025.
\bibitem{ye2025fitprune} W. Ye, Q. Wu, W. Lin, and Y. Zhou, ``Fit and Prune: Fast and training-free visual token pruning for multi-modal large language models,'' in \emph{Proc. AAAI}, vol. 39, no. 21, pp. 22128--22136, 2025.
\bibitem{guo2025crop} J. Guo, F. Zhai, P. Jian, Q. Wei, and Y. Zhou, ``CROP: Contextual region-oriented visual token pruning,'' in \emph{Proc. EMNLP}, pp. 9756--9772, 2025.
\bibitem{li2025fcot} J. Li, J. Fan, F. Tang, G. Huang, S. Zhu, S. Liu, N. Xie, W. Liu, and Y. Liao, ``FCoT-VL: Advancing text-oriented large vision-language models with efficient visual token compression,'' arXiv:2502.18512, 2025.
\bibitem{liu2025textguided} B. Liu, L. Qiao, Y. Wang, Z. Gao, Y. Ma, K. Ying, and T. Qin, ``Text-guided token communication for wireless image transmission,'' arXiv:2507.05781, 2025.
\bibitem{devoto2025adaptive} M. Devoto, J. Pomponi, M. Merluzzi, P. D. Lorenzo, and S. Scardapane, ``Adaptive semantic token communication for Transformer-based edge inference,'' arXiv:2505.17604, 2025.
\bibitem{li2026unisc} B. Li, X. Yang, S. Duan, and N. Wang, ``Toward universal semantic communication via matchable semantic subspace transmission,'' \emph{IEEE Trans. Image Process.}, vol. 35, pp. 5003--5016, 2026, doi: 10.1109/TIP.2026.3690331.
\bibitem{deniz2026adsc} O. F. Deniz, R. Mao, R. Li, Y. Tian, and L. Khan, ``Vision token reduction via attention-driven self-compression for efficient multimodal large language models,'' arXiv:2602.12618, 2026.
\bibitem{wan2026rtprune} B. Wan, Y. Feng, Z. Tang, W. Huang, Y. Zeng, J. Wang, and T. Liu, ``RTPrune: Reading-twice inspired token pruning for efficient DeepSeek-OCR inference,'' in \emph{Proc. ICML}, vol. 306, 2026.
\bibitem{chen2026energy} H. Chen and J. He, ``Energy-driven adaptive visual token pruning for efficient vision-language models,'' arXiv:2603.05950, 2026.
\bibitem{gu2026robust} S. Gu, J. Cui, W. Hu, Z. Shi, Z. Hu, and R. Hong, ``Visual token compression enhances robustness of MLLMs,'' arXiv:2607.22716, 2026.
\bibitem{wireless2026lic} M. Naseri, P. Ashtari, M. Seif, E. De Poorter, H. V. Poor, and A. Shahid, ``Deep learning-based image compression for wireless communications: Impacts on robustness, throughput, and latency,'' \emph{npj Wireless Technology}, vol. 2, art. no. 14, 2026, doi: 10.1038/s44459-025-00019-6.
\bibitem{hoydis2022sionna} J. Hoydis, S. Cammerer, F. Ait Aoudia, A. Vem, N. Binder, G. Marcus, and A. Keller, ``Sionna: An open-source library for next-generation physical layer research,'' arXiv:2203.11854, 2022. [Online]. Available: \url{https://nvlabs.github.io/sionna/}
\bibitem{llamagen2024} P. Sun, Y. Jiang, S. Chen, S. Zhang, B. Peng, P. Luo, and Z. Yuan, ``Autoregressive model beats diffusion: Llama for scalable image generation,'' arXiv:2406.06525, 2024. [Online]. Available: \url{https://github.com/FoundationVision/LlamaGen}
\bibitem{kodak24} Eastman Kodak Company, ``Kodak lossless true color image suite,'' [Online]. Available: \url{https://r0k.us/graphics/kodak/}. Accessed: Aug. 16, 2026.
\end{thebibliography}
\end{document}